\def\hideacknowledgments{}
\def\preprintcopy{}
\documentclass[cameraready]{Interspeech}
\ifdefined\XeTeXgenerateactualtext
\fi
\usepackage{fontspec}
\newfontfamily\banglafont{NotoSerifBengali-Regular}[
  Script=Bengali,
  Path=./,
  Extension=.ttf,
  Renderer=HarfBuzz
]

\normalfont
\newcommand{\bn}[1]{{\banglafont #1}}

\usepackage{array}
\usetikzlibrary{positioning,arrows.meta}

\newcommand{\dsname}{\textit{BanglaKontho}\xspace}

\title{BanglaKontho: Closing the Long-Form Gap in Bangla Text-to-Speech}

\author[affiliation={1}, orcid=0009-0000-2297-6037]{Mizbaul Haque}{Maruf}

\address{
    $^1$ Vivasoft Limited, Dhaka, Bangladesh
}

\email{mizbaul.haque@vivasoftltd.com}


\ifdefined\acceptancebanner
  \usepackage{eso-pic}
  \AddToShipoutPictureFG{%
    \ifnum\value{page}=1\relax
      \AtPageUpperLeft{%
        \raisebox{\dimexpr-12mm+\voffset\relax}{%
          \makebox[\paperwidth][c]{\small\acceptancebanner}%
        }%
      }%
    \fi
  }
\fi

\ifdefined\hideresourcelinks
  \newcommand{\resourcefootnote}{}
\else
  \newcommand{\resourcefootnote}{%
    \footnote{\raggedright Dataset and normalizer: \url{https://github.com/mizba-grad/BanglaKontho}\par}%
  }
\fi

\begin{document}

\maketitle
% Preprint copies only (build.sh defines \preprintcopy): the class writes the
% PDF Author field as "Mizbaul Haque {Maruf} {1};", which reference managers
% import verbatim. Must come after \maketitle, where the class sets it.
\ifdefined\preprintcopy
  \hypersetup{pdfauthor={Mizbaul Haque Maruf}}
\fi

% The abstract must exactly match the abstract entered into the submission system:
% 1000 characters, ASCII only, no citations, no LaTeX code.
\begin{abstract}
    Bangla, the seventh most spoken language in the world, remains under-resourced for neural text-to-speech. Public Bangla speech corpora are dominated by short read-prompt utterances collected for speech recognition, leaving long-form prosody and consistent single-speaker narration uncovered. We present BanglaKontho, a single-speaker Bangla TTS corpus of 20 hours derived from professional audiobook recordings: 7,050 segmented utterances with verified transcripts at 24 kHz. We also release a reusable Bangla text normalizer covering Bangladeshi-style digit grouping, currency and date expressions, Danda punctuation and Unicode normalization, together with the full preprocessing pipeline. An MB-iSTFT-VITS baseline trained from scratch reaches 9.5\% WER and 4.46 naturalness MOS, against 16.0\% and 3.16 for the same architecture retrained on the 12-hour IndicTTS-Bn corpus. The corpus is released openly under CC BY-NC 4.0.
\end{abstract}

%==================================================================================
\section{Introduction}
\label{sec:intro}

Bangla (Bengali) is the seventh most spoken language in the world, with over 230 million speakers, yet text-to-speech (TTS) research for Bangla materially lags English and Mandarin.
The principal barrier is data: high-quality TTS training corpora for Bangla are scarce, fragmented across multiple small releases, and dominated by short read-prompt recordings originally collected for automatic speech recognition (ASR) \cite{kjartansson2018crowd,commonvoice}.
Modern end-to-end neural TTS architectures such as VITS \cite{vits} and its lightweight variants \cite{mbistft} require tens of hours of clean speech aligned with verified text in order to produce natural, intelligible output.
Existing Bangla resources are insufficient for this regime, particularly when prosody, sentence-level intonation, and long-form rhythm are required for downstream applications such as audiobook synthesis, screen readers, and conversational agents.

In this paper we describe \dsname, a single-speaker Bangla audiobook corpus of 20 hours with verified transcripts, % The {} terminates the control word so the following newline still yields a
% space -- without it the suppressed-footnote build runs "4.0." into "The".
released openly under CC~BY-NC~4.0.\resourcefootnote{}
The audiobook domain is deliberate: studio recordings by a single trained narrator over long-form material yield acoustic and prosodic patterns that complement the read-prompt corpora that dominate existing Bangla TTS releases.
Together with the audio-text pairs, we release the full preprocessing pipeline and an open-source Bangla text normalizer.

% samepage keeps the "Contributions." heading glued to its list: without it the
% heading is left stranded at the foot of column 1 while the items start column 2.
% The whole block is ~14 lines, so it always fits in a column and simply moves
% to the next one as a unit.
\begin{samepage}
\noindent\textbf{Contributions.}
\begin{enumerate}
    \item The first dedicated single-speaker Bangla audiobook TTS corpus with verified transcripts: 20~hours / 7{,}050 utterances at \SI{24}{\kilo\hertz}, openly released under CC~BY-NC~4.0 (Section~\ref{sec:dataset}).
    \item An open-source Bangla text normalizer with Bangladeshi-style digit grouping, lakh / koti scale, decimal and date handling, and a strict Bangla-Unicode-block mode (Section~\ref{sec:norm}).
    \item A reproducible end-to-end preprocessing pipeline and an MB-iSTFT-VITS baseline with MCD, WER, and naturalness MOS, benchmarked against the same architecture retrained on IndicTTS-Bn and against a normalizer-disabled ablation (Section~\ref{sec:baseline}).
\end{enumerate}
\end{samepage}

%==================================================================================
\section{Related work}
\label{sec:related}

\noindent\textbf{Bangla speech resources.}
The most widely cited Bangla speech corpus is the crowd-sourced read-prompt collection of Kjartansson et al.\ \cite{kjartansson2018crowd}, hosted on OpenSLR, containing roughly 200~hours from over 500 speakers.
Its many-speaker design makes it well-suited for ASR but less so for single-voice TTS, and utterance durations are typically short.
The IndicTTS project~\cite{indictts} released paired studio TTS recordings for major Indian languages including Bangla, with roughly 12~hours per voice; Kumar et al.\ \cite{kumar2023indictts} later trained open FastPitch + HiFi-GAN models on this database for 13 languages.
The Bangla release contains more than one voice; throughout this paper IndicTTS-Bn refers to the single 12-hour Bangla voice we use as a baseline, not to the full Bangla release.
The Bangla subset of Mozilla Common Voice~\cite{commonvoice} consists of validated short-form crowd-sourced utterances.
Shrutilipi~\cite{shrutilipi} mines 6{,}400+ hours of weakly aligned Indic speech (including Bangla) from broadcast audio, but transcript-alignment quality is variable and the corpus targets ASR rather than TTS.
None of these resources simultaneously offers (i)~a single trained speaker, (ii)~long-form audiobook prosody, (iii)~studio-clean audio, and (iv)~verified transcripts.
Table~\ref{tab:related} summarizes the resource landscape.

\noindent\textbf{Bangla TTS.}
Early work on Bangla TTS adopted statistical-parametric methods~\cite{gutkin2016bangla} on small read-prompt collections such as SHRUTI~\cite{shruti}.
More recent neural approaches have explored zero-shot voice cloning~\cite{ahmed2024bengali}, achieving high speaker similarity but relying on centralized fine-tuning that aggregates raw speech on a server, with attendant privacy and licensing risks.
To our knowledge, no prior work has released a single-speaker, long-form Bangla audiobook corpus.
Outside Bangla, recent end-to-end systems including VITS2~\cite{vits2}, Matcha-TTS~\cite{matchatts}, and StyleTTS~2~\cite{styletts2} have set new quality bars; we adopt a VITS-based architecture for direct comparability with prior Bangla work, but the corpus is architecture-agnostic.

\noindent\textbf{Text normalization for Indic scripts.}
Bangla text normalization is non-trivial because of (i)~Bangladeshi-style digit grouping and non-compositional number words in the range 20--99, (ii)~conjunct characters and decomposed vowel signs that are visually identical but Unicode-distinct, and (iii)~mixed Bangla / ASCII digits in modern web sources.
Gutkin et al.\ \cite{gutkin2016bangla} note the need for language-specific normalization but do not release a reusable tool; we address this gap.

\begin{table*}[t]
  \caption{Bangla speech corpora most commonly used as TTS data sources. \dsname fills the long-form, single-speaker audiobook slot. The IndicTTS-Bn row describes the single 12-hour Bangla voice used as our baseline, not the full Bangla release.}
  \label{tab:related}
  \centering
  \begin{tabular}{lrrlrl}
    \toprule
    \textbf{Corpus} & \textbf{Hours} & \textbf{Speakers} & \textbf{Domain} & \textbf{SR (kHz)} & \textbf{Access} \\
    \midrule
    OpenSLR Bengali \cite{kjartansson2018crowd}  & $\sim$200 & 505  & Read prompts          & 16 & Open \\
    IndicTTS-Bn \cite{indictts}                  & 12        & 1    & Read prompts (studio) & 48 & Open \\
    Common Voice Bn \cite{commonvoice}           & 60+       & many & Read prompts          & 48 & Open \\
    Shrutilipi (Bn portion) \cite{shrutilipi}    & 437       & many & Broadcast             & 16 & Open \\
    \midrule
    \dsname (this work)                          & \textbf{20} & \textbf{1} & \textbf{Audiobook} & \textbf{24} & Open (CC~BY-NC~4.0) \\
    \bottomrule
  \end{tabular}
\end{table*}

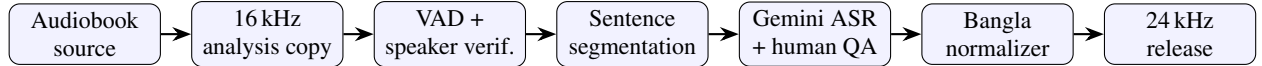
\begin{figure*}[t]
  \centering
  \begin{tikzpicture}[
      node distance=0.4cm and 0.4cm,
      block/.style={rectangle, draw, rounded corners, minimum width=2.0cm, minimum height=0.8cm, font=\small, align=center, fill=blue!5},
      arrow/.style={-{Stealth[length=2.5mm]}, thick}
  ]
  \node[block] (src)  {Audiobook \\ source};
  \node[block, right=of src]  (prep) {\SI{16}{\kilo\hertz} \\ analysis copy};
  \node[block, right=of prep] (vad)  {VAD + \\ speaker verif.};
  \node[block, right=of vad]  (seg)  {Sentence \\ segmentation};
  \node[block, right=of seg]  (asr)  {Gemini ASR \\ + human QA};
  \node[block, right=of asr]  (norm) {Bangla \\ normalizer};
  \node[block, right=of norm] (rel)  {\SI{24}{\kilo\hertz} \\ release};

  \draw[arrow] (src)  -- (prep);
  \draw[arrow] (prep) -- (vad);
  \draw[arrow] (vad)  -- (seg);
  \draw[arrow] (seg)  -- (asr);
  \draw[arrow] (asr)  -- (norm);
  \draw[arrow] (norm) -- (rel);
  \end{tikzpicture}
  \caption{End-to-end \dsname construction pipeline. A \SI{16}{\kilo\hertz} analysis copy of the source audiobook is filtered by voice activity detection and narrator verification and split at sentence boundaries; clips are transcribed by an ASR first pass followed by full human verification and normalized with the released Bangla text normalizer. The released audio is cut from the native-rate source at the resulting timings and resampled to \SI{24}{\kilo\hertz}.}
  \label{fig:pipeline}
\end{figure*}

%==================================================================================
\section{Dataset construction}
\label{sec:dataset}

Figure~\ref{fig:pipeline} summarizes the end-to-end pipeline. We describe each stage below.

\subsection{Audio acquisition and preprocessing}
Source audio was acquired in its native delivery format.
The source is a mastered audiobook delivery and therefore carries post-production processing such as dynamic-range compression, which we do not attempt to invert.
The pipeline maintains two parallel representations of every recording.
An \emph{analysis} copy at \SI{16}{\kilo\hertz}, 16-bit PCM, mono, with peak normalization, drives the segmentation and speaker-verification stages, which operate at the \SI{16}{\kilo\hertz} input rate expected by \texttt{pyannote.audio}; these stages yield only timing decisions.
The \emph{release} audio is then cut from the native-rate source at those timings and resampled to \textbf{\SI{24}{\kilo\hertz}}, the sampling rate of the LibriTTS corpus~\cite{libritts} and of universal neural vocoders trained on it, such as BigVGAN~\cite{bigvgan}; released clips therefore retain the full bandwidth of the source delivery.
Music intros/outros and other non-narrator material are removed automatically by the speaker-verification stage of Section~\ref{sec:spkverif}; the author spot-checked the retained and discarded regions by ear.

\subsection{Speaker verification}
\label{sec:spkverif}
Because the corpus is single-speaker by construction, the preprocessing challenge is not multi-speaker diarization but reliable removal of non-narrator audio (music intros / outros, chapter-announcement voices distinct from the narrator, occasional guest readings).
We use \texttt{pyannote.audio} 3.1 \cite{plaquet2023pyannote} for voice activity detection and segment-level speaker embeddings (ECAPA-TDNN \cite{ecapa}), then verify each speech segment against a manually enrolled narrator reference via cosine similarity.
Segments below a calibrated threshold are discarded.

\subsection{Segmentation}
Segmentation is guided by VAD-derived silence boundaries, using pause duration to favor cuts at likely sentence ends.
Clips shorter than \SI{1.0}{\second} or longer than \SI{20}{\second} are dropped; the operating target range is 2--\SI{15}{\second}, the standard window for most TTS.

\subsection{Transcription and manual verification}
\label{sec:transcription}
A first-pass transcript is produced by Google Gemini 2.0 Flash-Lite~\cite{gemini} for each clip, exploiting its broad Indic-language coverage.
On a randomly sampled 500-clip subset hand-transcribed by the author, Gemini achieved 2.7\% WER against the gold reference, which we treat as a strong but imperfect starting point rather than ground truth.
Two volunteer native Bangla speakers then independently reviewed 100\% of clips for word-level lexical accuracy, phoneme-relevant punctuation, and audio defects missed by the automated stage; word-level inter-annotator agreement was Cohen's $\kappa = 0.91$ on a 500-clip overlap subset, and remaining disagreements were resolved by discussion.

\subsection{Bangla text normalization}
\label{sec:norm}

We release \texttt{bangla\_normalize}, a stand-alone Python module used to clean every transcript in the corpus.
The pipeline is intended to be reusable across Bangla TTS / ASR projects and is byte-equivalent to the cleaner shipped in the training filelists.
\begin{enumerate}
    \item \textbf{Unicode NFC.} Decomposed vowel signs (e.g., \bn{ে}\,+\,\bn{া}~$\to$~\bn{ো}) are composed so that each visible grapheme corresponds to a single codepoint.
    \item \textbf{Symbol-to-word.} Currency symbols (\bn{৳}, \pounds, \texteuro, \bn{₹}), percent, ampersand, and other ASCII symbols are rewritten to their Bangla word equivalents.
    % \mbox: XeLaTeX (arXiv build) would otherwise break the line after the en-dash.
    \item \textbf{Numeric expansion} over both Bangla (\bn{০}--\bn{৯}) and ASCII \mbox{(0--9)} digits.
        The normalizer hard-codes the 80 unique Bangla compound words for 20--99, since Bangla number words in this range are not productively compositional (26 is \emph{chhabbish}, not \emph{bish-chhoy}).
    \item \textbf{Bangladeshi digit grouping} on the lakh / koti scale to arbitrary magnitude
        (e.g., \bn{১,২৩,৪৫,৬৭৮}~$\to$ \emph{ek koti teish lakh poyetallish hajar chhaysh\^o attatt\^or}).
    \item \textbf{Decimals and dates}: \bn{১২.৫} $\to$ \emph{baro doshomik p\=anch}; \bn{২০২৪-০৪-২৩} $\to$ \emph{dui hajar chobbish ch\=ar teish}.
    \item \textbf{Danda/Daari punctuation}: repeated Danda/Daari are collapsed (\bn{।।।}~$\to$~\bn{।}); double-Danda/Daari is folded (\bn{॥}~$\to$~\bn{।}); spacing around the Danda/Daari is normalized.
    \item \textbf{Strict mode (optional)} restricts output to the Bangla Unicode block \mbox{U+0980--U+09EF} plus a safe-punctuation set, suitable for downstream G2P with a fixed symbol table.
\end{enumerate}

Example: \bn{আজ ১২.৫ টাকা ও} 50\% \bn{ছাড়}~$\to$~\bn{আজ বারো দশমিক পাঁচ টাকা ও পঞ্চাশ শতাংশ ছাড়}.

\subsection{Quality filtering and statistics}
After transcription and normalization we drop clips with residual non-speech tails, anomalous duration-to-text ratios, or verifier flags from Section~\ref{sec:transcription}.
Final statistics are reported in Table~\ref{tab:stats}.

\begin{table}[t]
  \caption{\dsname dataset statistics. The chapter-disjoint split (no chapter contributes utterances to more than one of train / val / test) prevents same-chapter prosodic leakage between training and evaluation.}
  \label{tab:stats}
  \centering
  \begin{tabular}{lr}
    \toprule
    \textbf{Property} & \textbf{Value} \\
    \midrule
    Total duration            & \SI{20}{\hour} \\
    Utterances                & 7{,}050 \\
    Speakers                  & 1 \\
    Source domain             & Audiobook \\
    Sampling rate (release)   & \SI{24000}{\hertz} \\
    Sampling rate (baseline)  & \SI{22050}{\hertz} \\
    Mean utterance duration   & $\sim$\SI{10.2}{\second} \\
    Train / val / test split  & 6{,}078 / 500 / 472 \\
    Split policy              & chapter-disjoint \\
    Storage (parquet shards)  & $\sim$\SI{3.5}{\giga\byte} \\
    \bottomrule
  \end{tabular}
\end{table}

%==================================================================================
\section{Baseline TTS and evaluation}
\label{sec:baseline}

To benchmark the corpus we train MB-iSTFT-VITS~\cite{mbistft} from scratch.
MB-iSTFT-VITS replaces the HiFi-GAN-style decoder~\cite{hifigan} of standard VITS~\cite{vits} with a multi-band iSTFT head and PQMF subband synthesis, yielding a lightweight, real-time-friendly model directly comparable to prior single-speaker VITS work.
Training and evaluation audio is prepared at \SI{22050}{\hertz}, the canonical reference configuration of MB-iSTFT-VITS, from the \SI{16}{\kilo\hertz} analysis copy of Section~\ref{sec:dataset} rather than from the \SI{24}{\kilo\hertz} release; the baseline is therefore trained on band-limited audio and, if anything, understates what the released clips support.
All evaluation is reported at \SI{22050}{\hertz}.
The architecture combines a Transformer text encoder, a posterior encoder, a stochastic duration predictor, a normalizing-flow prior, and a multi-band iSTFT decoder, with 34.7\,M parameters at our configuration (hidden / inter 192, filter 768, two attention heads, six layers; ResBlock-1 kernels $[3,7,11]$; upsample rates $[4,4]$; initial channel 512; four PQMF subbands).
Training uses batch size 128 with AdamW ($\beta_1{=}0.8, \beta_2{=}0.99$), learning rate $2{\times}10^{-4}$ with decay $0.999875$, segment size 8192 samples, and FP16 mixed precision.
All training and preprocessing was conducted on a single NVIDIA H100 SXM5 \SI{80}{\giga\byte} GPU.

\subsection{Evaluation protocol}
We hold out 472 utterances as a test split, drawn from chapters disjoint from those used for training or validation to avoid intra-chapter prosodic leakage that would inflate evaluation scores under a naive utterance-level random split.
We report:
(i)~\textbf{MCD}~\cite{mcd}, DTW-aligned 13-dim MFCC distance;
(ii)~\textbf{WER} computed by transcribing the synthesized audio with IndicWav2Vec~\cite{indicwav2vec}, a Bangla-capable ASR (generic multilingual models such as Whisper-base~\cite{whisper} produce uninformative scores on Bangla synthesis); and
(iii)~\textbf{naturalness MOS} on a 5-point scale, with 15 volunteer native Bangla raters scoring 50 utterances per system (balanced for length) following the ITU-T P.800 protocol~\cite{p800}, reported as mean~$\pm$~95\% confidence interval.
All MOS sessions used headphones in a quiet environment, with system identity hidden and clip order randomized per rater; Krippendorff's $\alpha = 0.74$ across raters.
As an ASR-floor calibration, IndicWav2Vec on the \emph{natural} test-split audio (i.e., human ground truth) yields 8\% WER, so our reported synthesis WERs should be read relative to this floor rather than to 0\%.
We further include a \textbf{ground-truth MOS} ceiling obtained by submitting the natural test-split audio to the same rating protocol.
We compare our model against an MB-iSTFT-VITS baseline that we retrained from scratch on the single-speaker IndicTTS-Bn dataset~\cite{indictts} (12~hours) using the same configuration as our model; we do not use any released checkpoint.

\subsection{Baseline results}
Table~\ref{tab:eval} compares our MB-iSTFT-VITS model trained on \dsname against the same architecture retrained on IndicTTS-Bn and against the natural-speech ceiling.
Our model achieves substantially lower WER (9.5\% vs.\ 16.0\%) and higher naturalness MOS (4.46 vs.\ 3.16), with comparable MCD (5.85 vs.\ 5.97), and approaches the natural-speech ceiling of 4.90 MOS / 8\% ASR-floor WER.
Because the data-size gap between \dsname and IndicTTS-Bn is modest (\SI{20}{\hour} vs.\ \SI{12}{\hour}, a factor of 1.67), the large WER and MOS gaps cannot be explained by scale alone; they are consistent with the domain difference (consistent long-form audiobook narration vs.\ short read-prompt recordings) being the principal driver.

\begin{table}[t]
  \caption{Comparison of our VITS-based model trained on \dsname against the same architecture trained on IndicTTS-Bn (\SI{12}{\hour}) and against natural speech. MCD is in dB and WER in \%, both lower-is-better; WER is computed with IndicWav2Vec on the synthesized audio. Naturalness MOS is averaged over 15 native Bangla raters following the ITU-T P.800 protocol~\cite{p800} and reported with a 95\% confidence interval.}
  \label{tab:eval}
  \centering
  \setlength{\tabcolsep}{4pt}
  \begin{tabular}{lccc}
    \toprule
    \textbf{System} & \textbf{MCD} $\downarrow$ & \textbf{WER} $\downarrow$ & \textbf{MOS} $\uparrow$ \\
    \midrule
    \dsname (ours)    & 5.85 & 9.5  & $4.46 \pm 0.06$ \\
    \dsname w/o norm. & 5.81 & 14.0 & $3.80 \pm 0.08$ \\
    IndicTTS-Bn       & 5.97 & 16.0 & $3.16 \pm 0.09$ \\
    \midrule
    Natural speech    & --   & 8.0  & $4.90 \pm 0.03$ \\
    \bottomrule
  \end{tabular}
\end{table}

\subsection{Normalizer ablation}
To isolate the contribution of \texttt{bangla\_normalize}, we retrained the same MB-iSTFT-VITS architecture on \dsname with the normalizer disabled (raw verified transcripts only).
The ablated model degrades to WER 14.0\% and naturalness MOS 3.80 (Table~\ref{tab:eval}); MCD is comparable (5.81).
The gap is concentrated on utterances containing numerals, dates, currency symbols, or mixed Bangla/ASCII digits (18\% of the test set), where the ablated model frequently mispronounces symbol-bearing tokens; on the symbol-free remainder, the gap narrows substantially.
This pattern confirms that Bangla-specific text normalization contributes a localized but consistent intelligibility gain, a contribution distinct from corpus scale.

%==================================================================================
\section{Limitations and ethical considerations}
\label{sec:limitations}

\noindent\textbf{Scope.}
The corpus is single-speaker by construction; multi-speaker extension is left to future work.
The audiobook domain shapes prosody toward narrative reading style, which differs from conversational TTS; for conversational use we recommend combining \dsname with a small in-domain adaptation set.

\noindent\textbf{License and release.}
The full 20-hour \dsname corpus is openly released under CC~BY-NC~4.0.
The license does not rest on the recordings' public availability online: the rights holder of the source recordings has authorized their redistribution under this license, and the narrator has consented to the release.
The corpus consists of isolated sentence-level clips drawn from multiple books and distributed in shuffled order, so it does not reproduce any book as continuous text.
The accompanying preprocessing pipeline, including the \texttt{bangla\_normalize} library, is released under a permissive open-source license.

\noindent\textbf{Data provenance and ethical use.}
The corpus is redistributed strictly for non-commercial academic research; commercial use, voice cloning for impersonation, fraud, and the synthesis of political content are explicitly prohibited by the accompanying data-use agreement that all downloaders must accept.
We operate a standing takedown process: any request from a rights holder of the source books will be honored and the corresponding clips removed from the release.
For additional provenance protection we recommend that downstream systems trained on \dsname apply inaudible neural watermarking to synthesized audio.

\noindent\textbf{Speaker anonymity.}
We omit the narrator's identity from the released metadata, but acknowledge that voice-based identification remains possible for anyone in possession of the audio; we make no cryptographic anonymity claim about the speech signal itself.

\noindent\textbf{ASR-based intelligibility.}
WER depends on the Bangla ASR's own accuracy, which is improving rapidly but not error-free.
We accordingly report the ASR floor measured on natural speech (8\% WER, Table~\ref{tab:eval}) and read all synthesis WERs relative to that floor rather than to 0\%.

%==================================================================================
\section{Conclusion}
\label{sec:conclusion}

We release \dsname, a 20-hour single-speaker Bangla audiobook corpus built from professional recordings, together with a reusable Bangla text normalizer and an MB-iSTFT-VITS baseline.
The resource addresses a documented gap in long-form Bangla speech data and provides ready-to-use tooling for the broader Indic-TTS community.
Future work will (i)~extend to multiple narrators, (ii)~develop a Bangla-specific CMOS / preference-test benchmark, and (iii)~evaluate downstream applications such as screen readers and audiobook synthesis.

%==================================================================================
% Acknowledgments must not appear in the review version, and are dropped from the
% preprint copy, where build.sh defines \hideacknowledgments.
\ifcameraready
\ifdefined\hideacknowledgments\else
\section{Acknowledgments}
We thank the volunteer native Bangla speakers who verified the transcripts and
participated in the listening tests.
\fi
\fi

%==================================================================================
\section{Generative AI Use Disclosure}

Generative AI was used in two clearly delimited roles, both under full author
responsibility. First, as part of the described method, Google Gemini 2.0
Flash-Lite~\cite{gemini} produced the first-pass transcript of every clip
(Section~\ref{sec:transcription}); no machine-generated transcript entered the
release without independent verification by two native Bangla speakers, and the
measured quality of that first pass (2.7\% WER on a hand-transcribed subset) is
reported in the paper. Second, generative AI assistants were used for language
editing and formatting of the manuscript. No part of the scientific content,
experimental design, analysis, or conclusions was produced by a generative AI
tool, and the author takes full responsibility for the content of this paper.

%==================================================================================
\bibliographystyle{IEEEtran}
\bibliography{refs}

\end{document}